\documentclass[runningheads]{llncs}
\usepackage{makecell} 
\usepackage{array}    
\usepackage[T1]{fontenc}
\usepackage{graphicx}
\usepackage[colorlinks=true, allcolors=blue]{hyperref}

\hypersetup{hypertexnames=false}
\begin{document}
\title{Diminishing Returns in Self-Supervised Learning}
%
%
\author{
Oli Bridge$^{*}$ \quad
Huey Sun$^{*}$ \quad
Botond Branyicskai{-}Nagy$^{*}$ \quad
Charles D'Ornano$^{*}$ \quad
Shomit Basu$^{*}$ \\
}
\authorrunning{Bridge et al.}
%
\institute{University College London, London UK 
\email{ucabhss@ucl.ac.uk}\\
\url{https://www.ucl.ac.uk/computer-science/} \\
\bigskip
\textit{$^{*}$Equal contribution}}
\titlerunning{Diminishing Returns in Self-Supervised Learning}
\maketitle              
\begin{abstract}
Transformer-based architectures have become a dominant paradigm in vision and language, but their success is often attributed to large model capacity and massive training data. In this work, we examine how self-supervised pre-training, intermediate fine-tuning, and downstream fine-tuning interact in a low-capacity regime, using a 5M-parameter Vision Transformer for semantic segmentation.
Across multiple data scales, we find that masked image modeling pre-training and downstream fine-tuning reliably improve performance, but with clear diminishing returns as supervision increases. In contrast, inserting an intermediate classification fine-tuning stage consistently degrades downstream performance, with the largest drops occurring precisely where pre-training is most effective.
Through an analysis of patch-level representation geometry, we show that classification-based intermediate supervision actively interferes with representations learned during pre-training by collapsing spatial structure critical for dense prediction. These results indicate that, in small models, the geometry of supervision matters more than the number of training stages: misaligned intermediate objectives can negate the benefits of pre-training rather than amplify them.
\end{abstract}

\section{Introduction}

\subsection{Background}
Deep learning has driven major advances in computer vision, from image classification and object detection to semantic segmentation. However, many practical settings still face a scarcity of high-quality labeled data. Self-supervised learning (SSL) addresses this by pre-training models on large collections of unlabeled images and then fine-tuning them for a supervised downstream task, using pre-training to learn transferable representations.

Vision Transformers (ViTs) are a compelling architecture for SSL because they scale effectively with data and compute, representing images as sequences of patch tokens processed by self-attention \cite{maskedauto}. 
While prior work has demonstrated the benefits of SSL for Vision Transformers at scale \cite{ssl-vit}, it remains unclear how these methods behave when constrained to small models and limited data. In this work, we use a small 5M-parameter ViT (ViNy) to systematically examine how the amount of pre-training and fine-tuning data affects downstream semantic segmentation performance.

\begin{figure}[!t]
\centering
\includegraphics[width=.9\textwidth]{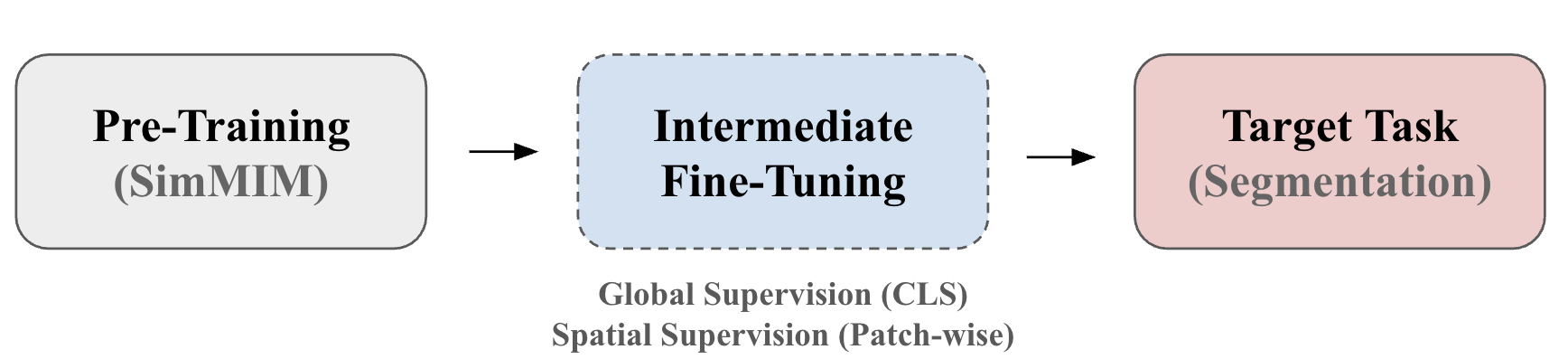}
\caption{Training pipeline for ViNy. The optional intermediate fine-tuning stage uses global (classification-based) or spatial (patch-aligned) supervision to isolate how the geometry of supervision affects transfer in low-capacity Vision Transformers.}
\label{fig:flow-pipeline}
\end{figure}

For pre-training, we use masked image modeling (MIM), where a model reconstructs masked portions of an input image \cite{genpix}. We adopt SimMIM \cite{simmim}, which masks image patches and predicts missing pixel values with a lightweight decoder, and pre-train on ImageNet-1K \cite{imagenet}. We also study intermediate fine-tuning \cite{stilts}, an additional supervised training stage performed before downstream fine-tuning that is commonly used to improve transfer in large models, but whose effect on dense prediction in low-capacity Vision Transformers remains poorly understood.

Overall, we evaluate how three stages—SimMIM pre-training, intermediate fine-tuning, and downstream segmentation fine-tuning—interact across data regimes, and when intermediate supervision helps versus harms transfer (Figure~\ref{fig:flow-pipeline}). We show that classification-based intermediate fine-tuning actively destroys spatial structure in small Vision Transformers, negating the benefits of self-supervised pre-training.

\subsection{Related Literature}

Intermediate fine-tuning has been extensively studied in natural language processing as a mechanism for improving transfer in pre-trained models. In NLP, self-supervised pre-training underlies models such as BERT \cite{bert}, while supervised intermediate fine-tuning on data-rich tasks \cite{stilts} was initially thought to improve downstream performance by transferring general skills.

Subsequent work has shown that this interpretation is incomplete. Chang and Lu \cite{rethinking-stilts} demonstrate that the effectiveness of intermediate fine-tuning depends on factors such as dataset size, task format, and alignment between training stages, and that intermediate objectives can influence downstream performance even when they are not semantically related. These findings raise broader questions about how intermediate supervision reshapes learned representations.

In computer vision, pre-training and intermediate fine-tuning have also been widely explored. Masked modeling has been successfully adapted to images through methods such as BEiT \cite{bao2022beit} and SimMIM \cite{simmim}, and intermediate fine-tuning is commonly applied to improve downstream classification and dense prediction performance. Prior work typically uses intermediate objectives that are closely aligned with the final task, and has shown that effective pre-training does not require massive datasets, particularly in limited-capacity regimes \cite{largescale}.

In contrast to prior vision work that emphasizes aligned intermediate objectives, we study intermediate fine-tuning under intentional task mismatch in a low-capacity Vision Transformer. Rather than proposing a new training strategy, we investigate when intermediate fine-tuning fails, and use simple representation diagnostics to characterize how misaligned supervision alters patch-level feature structure. This shifts attention from whether intermediate fine-tuning improves performance to when and why it can be detrimental in small-model, low-compute regimes.

\section{Methodology}

\subsection{ViNy}
All of our experiments use a small ViT architecture with an input image size of 128, a patch size of 16, an embedding dimension of 128, a depth of 12, 8 attention heads, and an MLP dimension of 512. This configuration, which we refer to as ViNy (for "tiny ViT"), contains 4.8M parameters, representing an 18× reduction compared to ViT-Base's 86M parameters.

\begin{figure}[t]
\centering
\includegraphics[width=.9\textwidth]{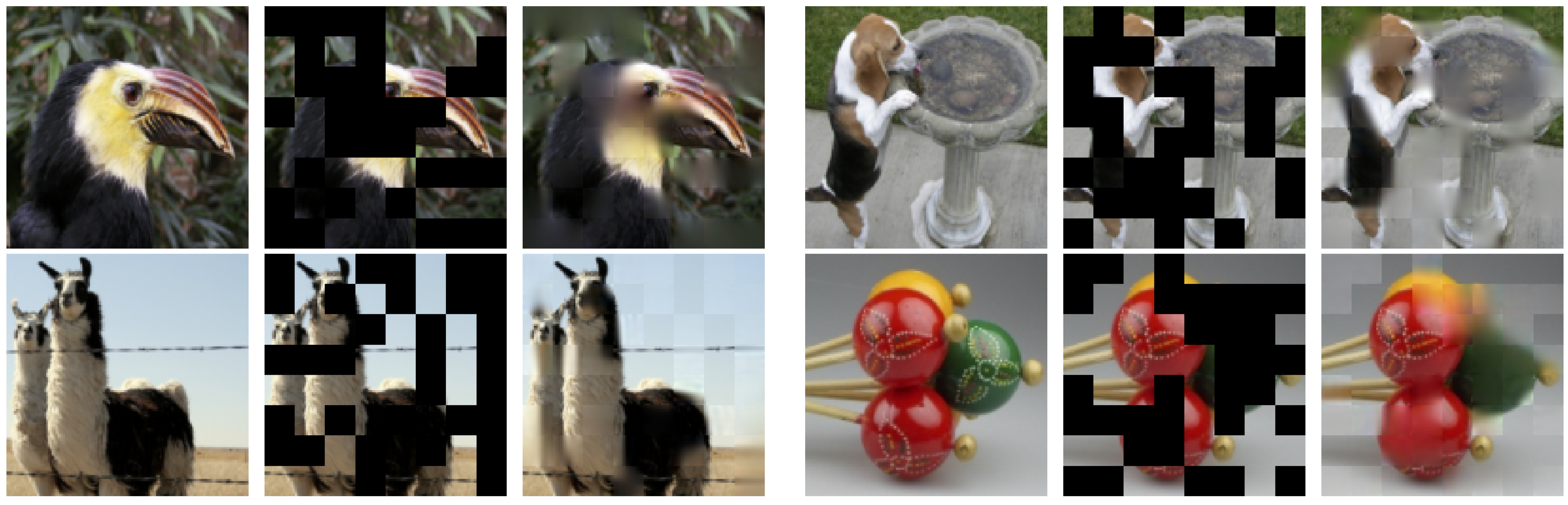}
\caption{Example SimMIM reconstructions on ImageNet-1K. Despite its small capacity (5M parameters), ViNy learns non-trivial patch-level structure during masked image modeling, providing a meaningful initialization for downstream tasks.}
\label{fig:simmim}
\end{figure}

\subsection{Pre-training with SimMIM}

We perform unsupervised pre-training using SimMIM \cite{simmim}, which randomly masks image patches and trains the model to reconstruct masked pixels using an L1 loss. We adopt a masking ratio of 0.5 and pre-train on ImageNet-1K \cite{imagenet}, varying the number of pre-training examples between 0, 45k, 100k, and 200k. Optimization details are provided in the appendix in  Table~\ref{tab:optimizer-config}.

\subsection{Intermediate Fine-tuning}

Unlike prior work that applies intermediate objectives aligned with the final task, we intentionally introduce a mismatch between intermediate and downstream supervision. Specifically, we perform intermediate fine-tuning on the Intel Image Classification dataset \cite{intel}, which contains approximately 14{,}000 images across six natural scene categories (e.g., buildings, forests, mountains). During this stage, ViNy is equipped with a $128\times6$ classification head, which is discarded before downstream segmentation fine-tuning. This design isolates the effect of global classification supervision on representations later used for dense prediction.

While a more closely aligned intermediate task would likely improve downstream accuracy, our goal is not to optimize performance but to characterize how misaligned intermediate supervision reshapes representations and affects transfer.

\subsection{Downstream Segmentation Fine-tuning}

The downstream task is semantic segmentation on the Oxford-IIIT Pet dataset \cite{pets}, using tri-map labels for foreground, background, and ambiguous pixels. We vary the number of training examples between 250 and 6000 and evaluate on 1000 held-out test images using a fixed random seed. ViNy is equipped with a patch-level prediction head mapping each patch embedding to a $16\times16\times3$ output. We report pixel accuracy and mean intersection-over-union (mIoU). Example segmentation outputs are shown in Appendix Figure~\ref{fig:seg-ex}. Code to reproduce all experiments is at \url{https://github.com/itshuey/masked-image-modelling}.



\section{Experiments}

First, we conduct an experiment to determine how the baseline ViNy model performs as we increase the number of fine-tuning examples, evaluating accuracy and mIoU on 1000 held-out test examples (Table \ref{tab:baseline-performance}).

\begin{table}[ht]
\centering
\caption{Baseline ViNy performance over 3 runs by number of fine-tuning examples}
\label{tab:baseline-performance}
\begin{tabular}{|c|@{\hspace{.5em}}c@{\hspace{.5em}}|@{\hspace{.5em}}c@{\hspace{.5em}}|}
\hline
\textbf{\# Finetune} & \textbf{Accuracy} & \textbf{mIoU} \\ \hline\hline
250 & 70.33 ± 0.89\% & 43.73 ± 1.02\% \\ 
500 & 73.71 ± 0.08\% & 47.41 ± 0.21\% \\
1000 & 77.81 ± 0.56\% & 53.28 ± 0.43\% \\
2000 & 80.03 ± 0.42\% & 56.87 ± 0.51\% \\
4000 & 84.16 ± 0.06\% & 62.99 ± 0.16\% \\
6000 & 85.60 ± 0.03\% & 65.07 ± 0.01\% \\
\hline
\end{tabular}
\end{table}

We then repeat this experiment for ViNy models pre-trained using SimMIM with a varying number of training examples from ImageNet 1-K. The results are visualized in Figure \ref{fig:perf-trend} and listed in Table \ref{tab:pt-performance} in the appendix.


\begin{figure}[ht]
\caption{Downstream segmentation mIoU as a function of pre-training scale and downstream fine-tuning data.}
\label{fig:perf-trend}
\centering
\includegraphics[width=.8\textwidth]{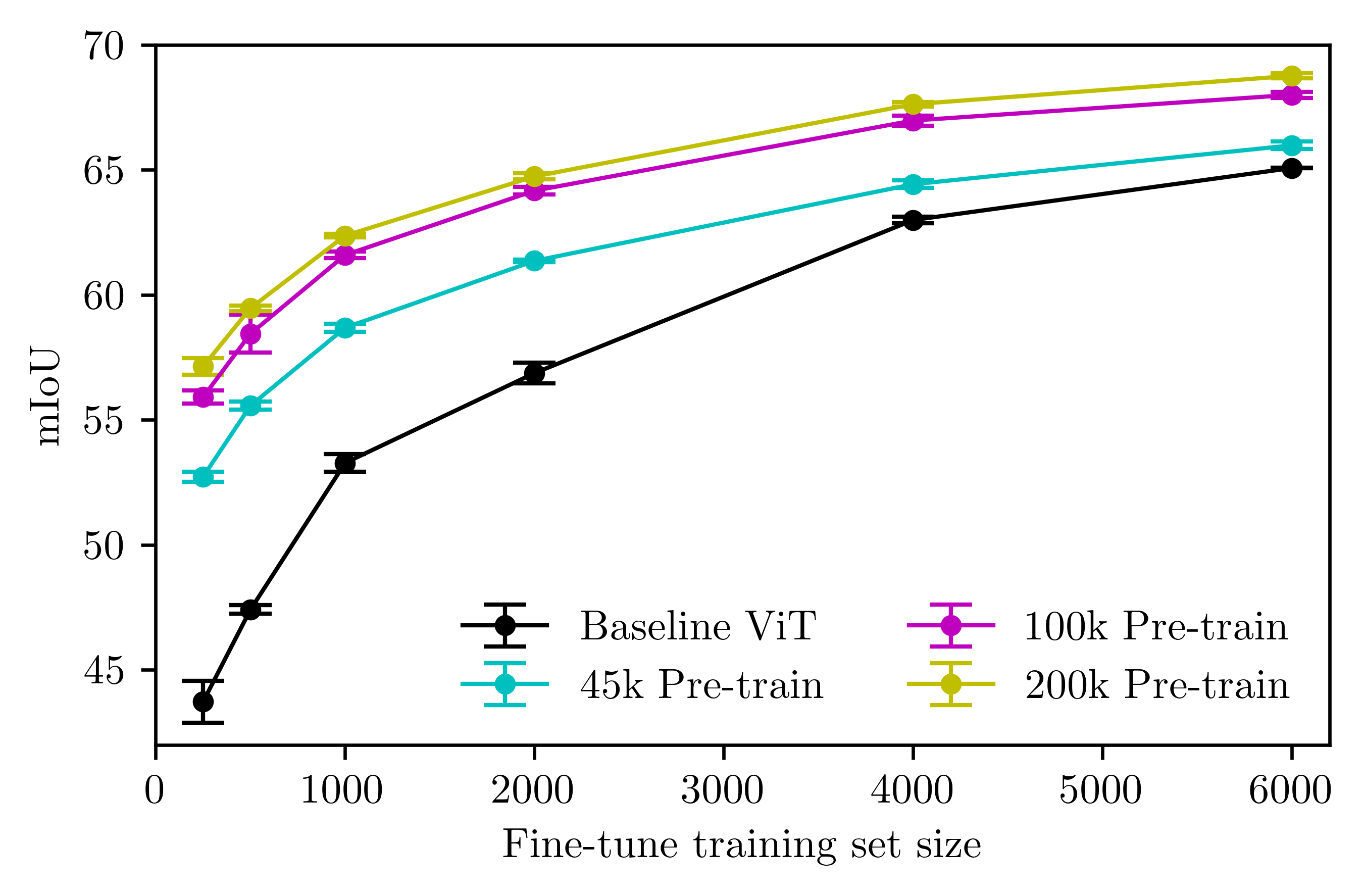}
\end{figure}

We then repeat this experiment with the inclusion of an intermediate fine-tuning stage. The impact of this additional supervision is visualized in Figure \ref{fig:it-trend} and the data is listed in Table \ref{tab:full-it-performance} in the appendix. As intermediate fine-tuning worsens performance, we show the paired results in terms of negative mIoU. All experiments are averaged over three runs with fixed data splits to control for variance.

\begin{figure}[!t]
\caption{Change in downstream mIoU induced by adding an intermediate classification fine-tuning stage. All models receive identical intermediate supervision on Intel Image Classification between pre-training and downstream segmentation fine-tuning.}
\label{fig:it-trend}
\centering
\includegraphics[width=.8\textwidth]{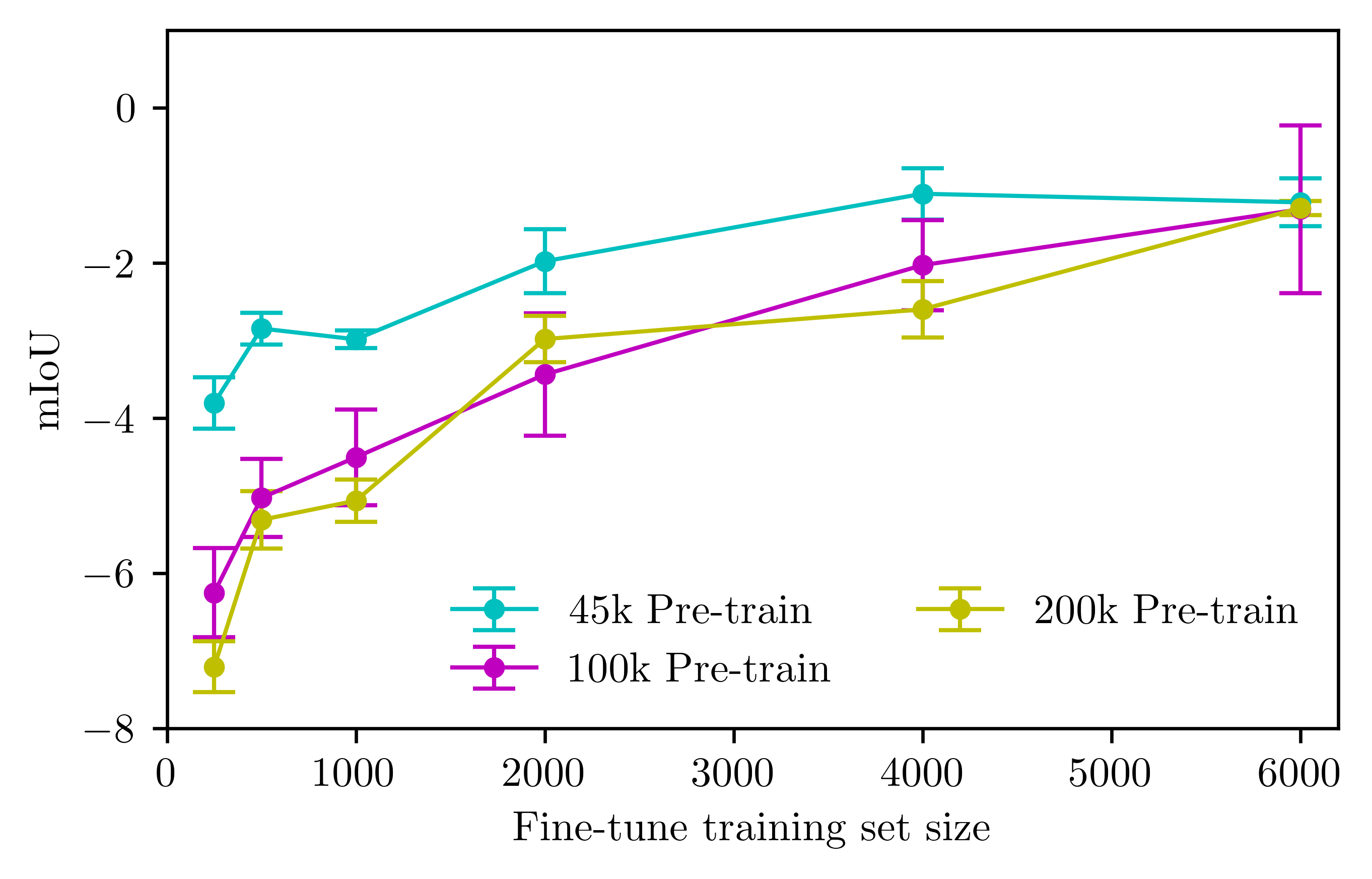}
\end{figure}
\section{Results}

\subsection{Performance effects of pre-training and intermediate supervision}

Pre-training substantially improves downstream segmentation performance in low-data regimes, yielding gains of up to approximately 13 mIoU points, but exhibits clear diminishing returns as the amount of downstream supervision increases (Figure~\ref{fig:perf-trend}). Increasing the scale of pre-training beyond 100k ImageNet examples produces smaller marginal improvements, which for a model as small as ViNy (5M parameters) likely reflects a capacity bottleneck rather than insufficient data.

In contrast, adding an intermediate classification fine-tuning stage consistently degrades downstream segmentation performance across all configurations (Figure~\ref{fig:it-trend}). The degradation is largest precisely in regimes where SimMIM pre-training is most effective, indicating that intermediate supervision interferes with—rather than complements—representations learned during self-supervised pre-training.

\subsection{Intermediate fine-tuning alters spatial representation geometry}

To better understand this degradation, we analyze how different types of supervision reshape the spatial geometry of patch-level representations. Because semantic segmentation depends on preserving fine-grained spatial structure, effective representations should be locally coherent within regions while still allowing sharp discontinuities at object boundaries. Accordingly, patch embeddings should exhibit higher similarity for nearby patches on average, with similarity decreasing as spatial distance increases, while permitting non-local similarity between semantically related regions.

\begin{figure}[!t]
\caption{Patch similarity as a function of spatial distance on the ViT patch grid. Curves show cosine similarity between patch embeddings at increasing Manhattan distances.}
\label{fig:patch-sim}
\centering
\vspace{10pt}
\includegraphics[width=.7\textwidth]{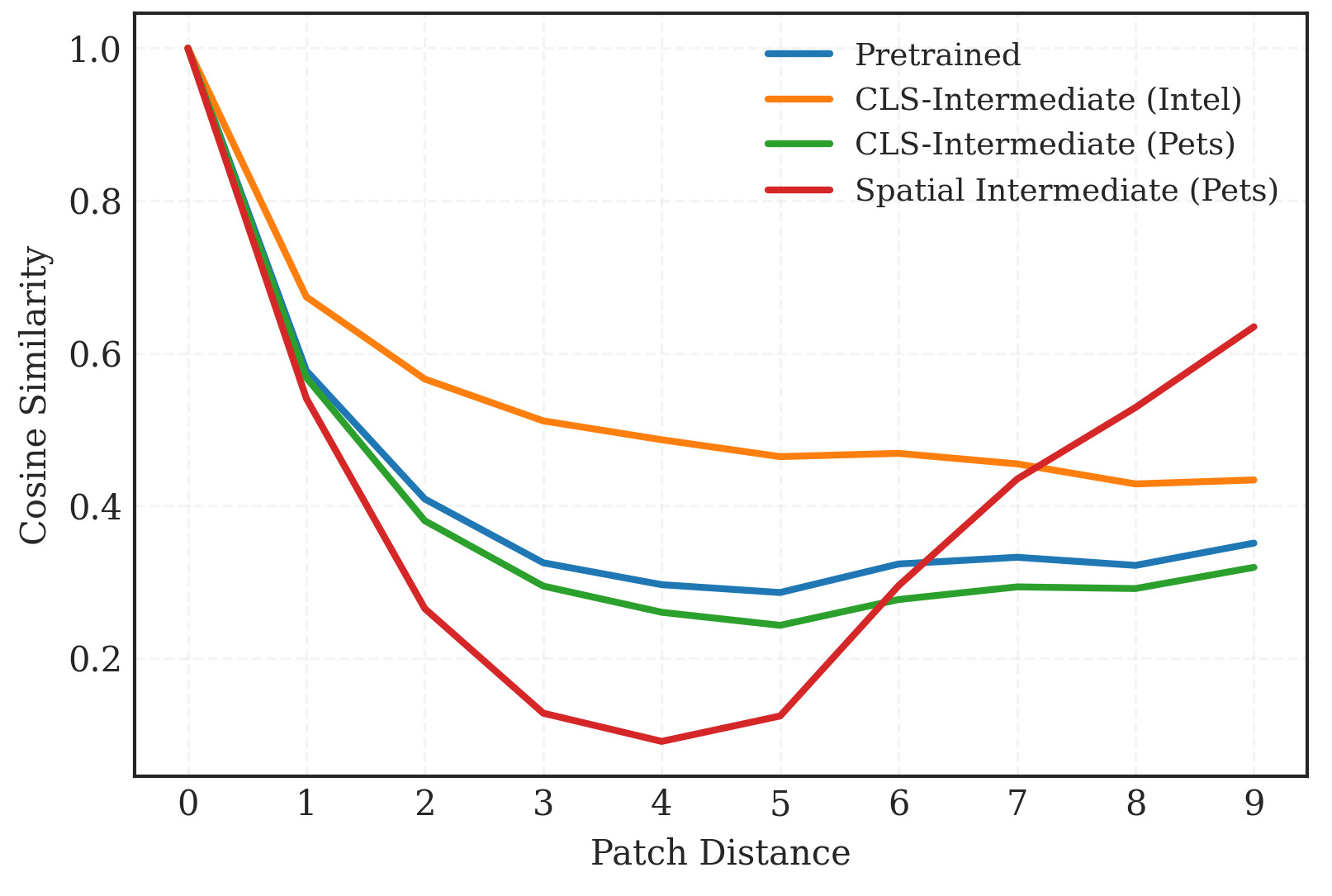}
\vspace{-6pt}
\end{figure}

\begin{table}[!t]
\centering
\caption{Patch-level representation diagnostics across training regimes. Patch Probe Acc measures linear separability of patch features for segmentation classes.}
\label{tab:patch-diagnostics}
\setlength{\tabcolsep}{4pt} 
\begin{tabular}{|l|c|c|c|}
\hline
\textbf{Model} & \makecell[c]{\textbf{Within-Image} \\ \textbf{Variance}} & \makecell[c]{\textbf{Across-Image} \\ \textbf{Variance}} & \makecell[c]{\textbf{Patch Probe} \\ \textbf{Acc (\%)}} \\ \hline\hline
Pretrained (SimMIM)      & 0.10 & 0.02 & 79.5 \\ 
CLS-Intermediate (Intel) & 0.18 & 0.11 & 76.1 \\ 
CLS-Intermediate (Pets)  & 0.58 & 0.29 & 78.6  \\ 
Spatial Intermediate     & 0.32 & 0.06 & \textbf{93.3} \\ 
\hline
\end{tabular}
\end{table}

To isolate the role of supervision geometry, we include \emph{spatial intermediate} supervision as a diagnostic control. In this setting, the model is trained on the same dataset as the classification-based intermediate task, but with patch-aligned, spatially local targets rather than a global classification objective, allowing us to attribute observed differences to supervision geometry rather than dataset shift or supervision strength.

\begin{figure}[!t]
\caption{Emergent patch-level similarity structure under different training regimes. Each heatmap shows cosine similarity between a reference patch (red square) and all other patches in the image. Classification-based intermediate fine-tuning produces spatially diffuse, globally similar representations, while spatially aligned supervision preserves localized structure consistent with dense prediction.}
\label{fig:patch-structure}
\centering
\vspace{2pt}
\includegraphics[width=.9\textwidth]{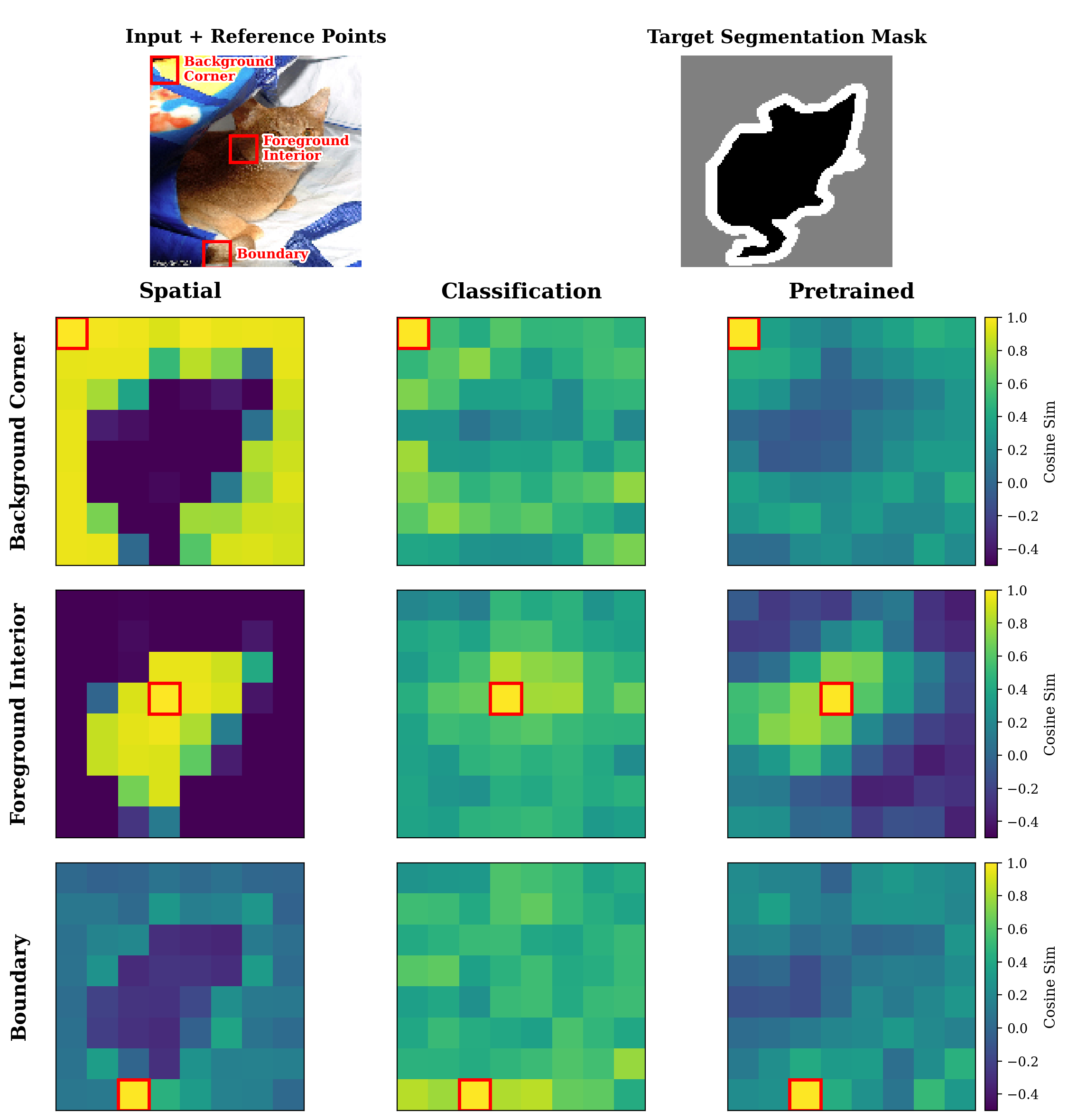}
\end{figure}

Figure~\ref{fig:patch-sim} plots cosine similarity between patch tokens as a function of spatial distance on the ViT patch grid. Across most regimes—including SimMIM pre-training and classification-based intermediate fine-tuning on Intel—similarity declines gradually with distance. Classification-based intermediate fine-tuning on the downstream dataset (CLS-Pets) increases overall similarity but does not substantially alter the shape of the similarity-distance curve, suggesting a dataset-specific effect rather than a fundamental geometric change.

At first glance, the similar distance-decay profiles for SimMIM pre-training and CLS-Intel raise a puzzle: if the curves are nearly identical, why does intermediate fine-tuning degrade downstream segmentation performance? The distinction lies not in the average similarity profile, but in the structure of similarity across individual patches.

Figure~\ref{fig:patch-structure} makes this concrete: each heatmap visualizes cosine similarity between a reference patch (red square) and all other patches in the image. Representations learned with SimMIM pre-training exhibit weak but discernible spatial structure, with coherent regions emerging amidst background noise. Classification-based intermediate fine-tuning preserves the overall similarity scale but substantially increases spatial noise, obscuring region-level organization even when average distance-dependent trends remain unchanged. The resulting segmentation predictions exhibit spatially unstructured errors (Appendix Figure~\ref{fig:seg-ex-bad}).

In contrast, spatially aligned intermediate supervision induces a qualitatively different representational structure. Nearby patches within the same region exhibit high similarity (local coherence), while distant patches corresponding to the same semantic class—most often background—also show elevated similarity (semantic grouping). Together, these complementary structures produce the characteristic U-shaped profile in Figure~\ref{fig:patch-sim}, where similarity is high at both short and long distances but drops at intermediate scales. This reflects semantically organized patch representations suitable for dense prediction, rather than the noisy, unstructured patterns produced by classification-based supervision.

These geometric differences translate directly into downstream utility. As shown in Table~\ref{tab:patch-diagnostics}, spatial intermediate supervision yields highly linearly separable patch features, achieving substantially higher probe accuracy than either pre-training alone or classification-based intermediate fine-tuning. In contrast, classification-based supervision increases variance without improving the usability of patch representations for dense prediction.

Together, these results show that in ViNy, intermediate fine-tuning reshapes patch-level representations in ways that directly affect downstream segmentation performance. Spatially aligned supervision sharpens and organizes patch representations, while classification-based supervision primarily injects noise and weakens the spatial structure learned during pre-training.

\section{Discussion}

Our results suggest that, in low-capacity Vision Transformers, the benefits of additional training stages depend critically on the geometry of supervision rather than the number of stages alone. While self-supervised pre-training provides a strong initialization for dense prediction, intermediate classification fine-tuning can systematically disrupt spatial structure and negate these gains.

A plausible explanation for this behavior lies in the geometry of the classification objective itself. In Vision Transformers, classification loss is mediated through a global [CLS] token, encouraging patch tokens to contribute to image-level discrimination rather than encode spatially local features. This incentive structure increases variance across images while weakening spatial organization within them (Table~\ref{tab:patch-diagnostics}), producing the spatially unstructured representations observed under classification-based intermediate fine-tuning. When representational capacity is limited and features are easily overwritten, this global pressure overwrites spatial structure learned during self-supervised pre-training rather than complementing it.

These findings suggest a practical guideline for low-compute regimes: prioritize downstream supervision, apply self-supervised pre-training where possible, and use intermediate fine-tuning only when its supervision structure aligns with the downstream task. In small models, task alignment matters more than the number of training stages, and indiscriminate intermediate fine-tuning can undo the benefits of pre-training rather than amplify them.
%
%
%
%
\newpage
\bibliography{references}

@INPROCEEDINGS{simmim,
  author={Xie, Zhenda and Zhang, Zheng and Cao, Yue and Lin, Yutong and Bao, Jianmin and Yao, Zhuliang and Dai, Qi and Hu, Han},
  booktitle={2022 IEEE/CVF Conference on Computer Vision and Pattern Recognition (CVPR)}, 
  title={SimMIM: a Simple Framework for Masked Image Modeling}, 
  year={2022},
  volume={},
  number={},
  pages={9643-9653},
  doi={10.1109/CVPR52688.2022.00943}}

@InProceedings{genpix,
  title = 	 {Generative Pretraining From Pixels},
  author =       {Chen, Mark and Radford, Alec and Child, Rewon and Wu, Jeffrey and Jun, Heewoo and Luan, David and Sutskever, Ilya},
  booktitle = 	 {Proceedings of the 37th International Conference on Machine Learning},
  pages = 	 {1691--1703},
  year = 	 {2020},
  editor = 	 {III, Hal Daumé and Singh, Aarti},
  volume = 	 {119},
  series = 	 {Proceedings of Machine Learning Research},
  month = 	 {13--18 Jul},
  publisher =    {PMLR},
  url = 	 {https://proceedings.mlr.press/v119/chen20s.html},
}

@INPROCEEDINGS{maskedauto,
  author={He, Kaiming and Chen, Xinlei and Xie, Saining and Li, Yanghao and Dollár, Piotr and Girshick, Ross},
  booktitle={2022 IEEE/CVF Conference on Computer Vision and Pattern Recognition (CVPR)}, 
  title={Masked Autoencoders Are Scalable Vision Learners}, 
  year={2022},
  volume={},
  number={},
  pages={15979-15988},
  doi={10.1109/CVPR52688.2022.01553}}

@inproceedings{
bao2022beit,
title={{BE}iT: {BERT} Pre-Training of Image Transformers},
author={Hangbo Bao and Li Dong and Songhao Piao and Furu Wei},
booktitle={International Conference on Learning Representations},
year={2022},
url={https://openreview.net/forum?id=p-BhZSz59o4}
}

@article{stilts,
  author       = {Jason Phang and Thibault F{\'{e}}vry and Samuel R. Bowman},
  title        = {Sentence Encoders on STILTs: Supplementary Training on Intermediate
                  Labeled-data Tasks},
  journal      = {CoRR},
  volume       = {abs/1811.01088},
  year         = {2018},
  url          = {http://arxiv.org/abs/1811.01088},
  eprinttype    = {arXiv},
  eprint       = {1811.01088},
  bibsource    = {dblp computer science bibliography, https://dblp.org}
}

@inproceedings{rethinking-stilts,
    title = "Rethinking Why Intermediate-Task Fine-Tuning Works",
    author = "Chang, Ting-Yun and Lu, Chi-Jen",
    editor = "Moens, Marie-Francine  and
      Huang, Xuanjing  and
      Specia, Lucia  and
      Yih, Scott Wen-tau",
    booktitle = "Findings of the Association for Computational Linguistics: EMNLP 2021",
    month = nov,
    year = "2021",
    address = "Punta Cana, Dominican Republic",
    publisher = "Association for Computational Linguistics",
    url = "https://aclanthology.org/2021.findings-emnlp.61",
    doi = "10.18653/v1/2021.findings-emnlp.61",
    pages = "706--713",
}

@article{largescale,
  title={Are Large-scale Datasets Necessary for Self-Supervised Pre-training?},
  author={Alaaeldin El-Nouby and Gautier Izacard and Hugo Touvron and Ivan Laptev and Herv{\'e} J{\'e}gou and Edouard Grave},
  journal={ArXiv},
  year={2021},
  volume={abs/2112.10740},
  url={https://api.semanticscholar.org/CorpusID:245334705}
}

@INPROCEEDINGS{ssl-vit,
  author={Chen, Xinlei and Xie, Saining and He, Kaiming},
  booktitle={2021 IEEE/CVF International Conference on Computer Vision (ICCV)}, 
  title={An Empirical Study of Training Self-Supervised Vision Transformers}, 
  year={2021},
  volume={},
  number={},
  pages={9620-9629},
  doi={10.1109/ICCV48922.2021.00950}}

@INPROCEEDINGS{pets,
  author={Parkhi, Omkar M and Vedaldi, Andrea and Zisserman, Andrew and Jawahar, C. V.},
  booktitle={2012 IEEE Conference on Computer Vision and Pattern Recognition}, 
  title={Cats and dogs}, 
  year={2012},
  volume={},
  number={},
  pages={3498-3505},
  doi={10.1109/CVPR.2012.6248092}}

@misc{intel,
    title={Intel Image Classification Dataset},
url = {https://www.kaggle.com/puneet6060/intel-image-classification}}

@inproceedings{bert,
    title = "{BERT}: Pre-training of Deep Bidirectional Transformers for Language Understanding",
    author = "Devlin, Jacob  and
      Chang, Ming-Wei  and
      Lee, Kenton  and
      Toutanova, Kristina",
    editor = "Burstein, Jill  and
      Doran, Christy  and
      Solorio, Thamar",
    booktitle = "Proceedings of the 2019 Conference of the North {A}merican Chapter of the Association for Computational Linguistics",
    month = jun,
    year = "2019",
    address = "Minneapolis, Minnesota",
    publisher = "Association for Computational Linguistics",
    url = "https://aclanthology.org/N19-1423",
    doi = "10.18653/v1/N19-1423",
    pages = "4171--4186",
}

@article{imagenet,
author = {Russakovsky, Olga and Deng, Jia and Su, Hao and Krause, Jonathan and Satheesh, Sanjeev and Ma, Sean and Huang, Zhiheng and Karpathy, Andrej and Khosla, Aditya and Bernstein, Michael and Berg, Alexander and Fei-Fei, Li},
year = {2014},
month = {09},
pages = {},
title = {ImageNet Large Scale Visual Recognition Challenge},
volume = {115},
journal = {International Journal of Computer Vision},
doi = {10.1007/s11263-015-0816-y}
}






\newpage
\section{Appendix}
\setcounter{table}{0}
\setcounter{figure}{0}
\renewcommand{\thetable}{A\arabic{table}}
\renewcommand{\thefigure}{A\arabic{figure}}

\begin{table}[ht]
\centering
\caption{Training Configurations for the Optimizer and Learning Rate Scheduler}
\label{tab:optimizer-config}
\begin{tabular}{|c|c|c|c|c|c|}
\hline
 && \multicolumn{2}{c|}{\textbf{AdamW Params}} & \multicolumn{2}{c|}{\textbf{MultiStepLR Params}} \\ \hline
                   \textbf{Phase} & \textbf{Epochs} & \textbf{Learning Rate} & \textbf{Weight Decay} & \textbf{Gamma} & \textbf{Milestones} \\ \hline
Pre-training       & 100 & $1 \times 10^{-4}$     & 0.05                  & 0.1           & \{50, 85\} \\ \hline 
Intermediate       & 200 & $8 \times 10^{-4}$     & 0                 & 0.1           & \{180, 190\} \\ \hline
Fine-tuning       & 100 & $2\times 10^{-3}$     & $1 \times 10^{-4} $                 & 0.5           & \{70, 90, 95\} \\ \hline
\end{tabular}
\end{table}

\begin{figure}[ht!]
\caption{A semantic segmentation example on the Oxford III Pets data set. The left image is the input, the center image is the ground truth, and the right image is our model's prediction}
\label{fig:seg-ex}
\vspace{2pt}
\centering
\includegraphics[width=.92\textwidth]{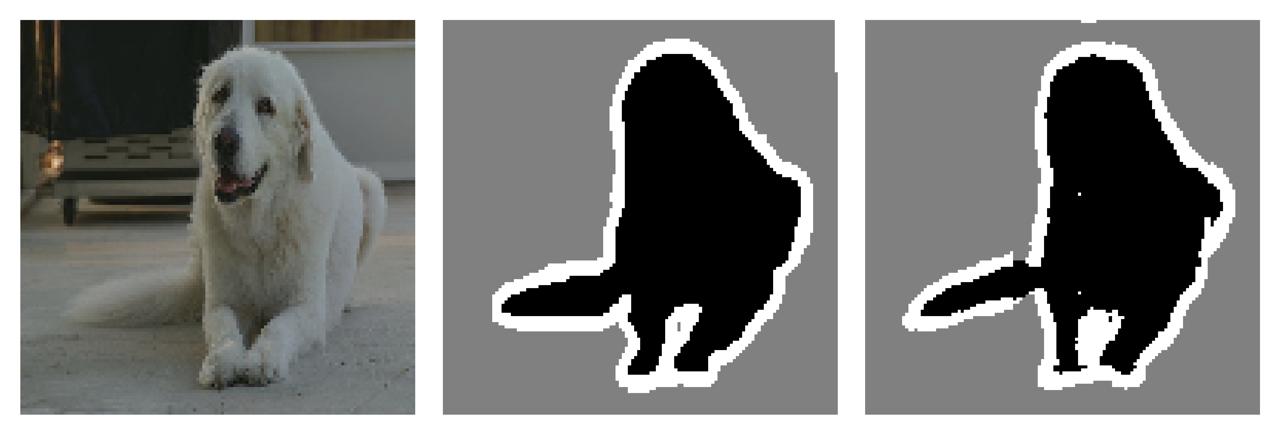}
\end{figure}

\begin{figure}[ht!]
\caption{Example segmentation failures after classification-based intermediate fine-tuning. Errors are spatially diffuse and unstructured, consistent with the loss of spatial locality observed in patch-level similarity analyses.}
\label{fig:seg-ex-bad}
\vspace{6pt}
\centering
\includegraphics[width=.92\textwidth]{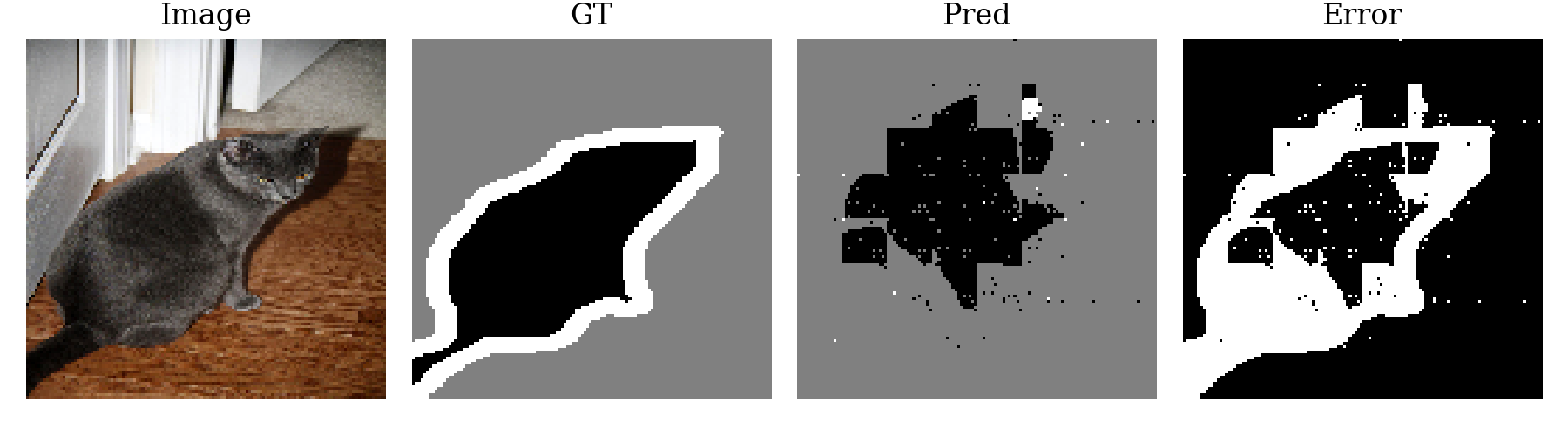}
\end{figure}

\begin{table}[ht]
\centering
\caption{Pre-trained ViNy test accuracy and mIoU across 3 runs by number of pre-training examples and number of fine-tuning examples}
\label{tab:pt-performance}
\begin{tabular}{|@{\hspace{.5em}}c@{\hspace{.5em}}|@{\hspace{.5em}}c@{\hspace{.5em}}|@{\hspace{.5em}}c@{\hspace{.5em}}|@{\hspace{.5em}}c@{\hspace{.5em}}|}
\hline
\textbf{\# Pre-train}&\textbf{\# Fine-tune} & \textbf{Accuracy} & \textbf{mIoU} \\ \hline\hline

45k & 250 & 77.44 ± 0.06\% & 52.73 ± 0.25\% \\
&500&79.67 ± 0.20\% & 55.57 ± 0.19\% \\
&1000&81.92 ± 0.20\% & 58.68 ± 0.20\% \\
&2000 &83.38 ± 0.03\% & 61.36 ± 0.05\% \\
&4000&85.41 ± 0.13\% & 64.42 ± 0.19\% \\
&6000&86.40 ± 0.11\% & 65.98 ± 0.19\% \\\hline

100k & 250 & 79.67 ± 0.35\% & 55.91 ± 0.32\% \\ 
&500 & 81.62 ± 0.62\% & 58.44 ± 0.93\% \\
&1000 & 83.60 ± 0.14\% & 61.59 ± 0.16\% \\
&2000 & 84.99 ± 0.18\% & 64.17 ± 0.18\% \\
&4000 & 86.81 ± 0.12\% & 66.97 ± 0.25\% \\
&6000 & 87.53 ± 0.08\% & 68.00 ± 0.14\% \\\hline

200k & 250 & 80.49 ± 0.18\% & 57.14 ± 0.42\% \\ 
&500 & 82.31 ± 0.04\% & 59.47 ± 0.13\% \\
&1000 & 84.17 ± 0.15\% & 62.37 ± 0.08\% \\
&2000 & 85.38 ± 0.11\% & 64.74 ± 0.14\% \\
&4000 & 87.23 ± 0.06\% & 67.62 ± 0.12\% \\
&6000 & 87.93 ± 0.04\% & 68.76 ± 0.13\%\\\hline

\end{tabular}
\end{table}

\begin{table}[ht]
\centering
\caption{Pre-trained and intermediate fine-tuned ViNy test accuracy and mIoU across 3 runs by number of pre-training examples and number of fine-tuning examples}
\label{tab:full-it-performance}
\begin{tabular}{|@{\hspace{.5em}}c@{\hspace{.5em}}|@{\hspace{.5em}}c@{\hspace{.5em}}|@{\hspace{.5em}}c@{\hspace{.5em}}|@{\hspace{.5em}}c@{\hspace{.5em}}|}
\hline
\textbf{\# Pre-train}&\textbf{\# Fine-tune} & \textbf{Accuracy} & \textbf{mIoU} \\ \hline\hline

45k & 250 & 74.65 ± 0.97\% & 48.92 ± 0.66\% \\
&500& 77.86 ± 0.32\% & 52.72 ± 0.39\% \\
&1000&79.80 ± 0.11\% & 55.70 ± 0.13\% \\
&2000 &82.08 ± 0.57\% & 59.38 ± 0.88\% \\
&4000&84.86 ± 0.28\% & 63.31 ± 0.68\% \\
&6000&85.70 ± 0.42\% & 64.76 ± 0.62\%\\\hline

100k & 250 & 75.72 ± 0.49\% & 50.20 ± 0.29\% \\ 
&500 & 78.59 ± 0.25\% & 53.85 ± 0.44\% \\
&1000 & 81.13 ± 0.19\% & 57.80 ± 0.10\% \\
&2000 & 83.77 ± 0.26\% & 61.62 ± 0.11\% \\
&4000 & 86.13 ± 0.11\% & 65.29 ± 0.61\% \\
&6000 & 87.32 ± 0.48\% & 67.79 ± 0.43\% \\\hline

200k & 250 & 75.83 ± 0.66\% & 49.93 ± 0.56\% \\ 
&500 & 78.49 ± 0.69\% & 54.15 ± 0.77\% \\
&1000 & 80.96 ± 0.26\% & 57.30 ± 0.57\% \\
&2000 & 83.96 ± 0.40\% & 61.75 ± 0.62\% \\
&4000 & 85.88 ± 0.48\% & 65.02 ± 0.77\% \\
&6000 & 87.18 ± 0.14\% & 67.47 ± 0.15\%\\\hline

\end{tabular}
\end{table}

\end{document}